\documentclass{article}

\usepackage[preprint]{neurips_2026}

\usepackage[utf8]{inputenc} 
\usepackage[T1]{fontenc}    
\usepackage{hyperref}       
\usepackage{url}            
\usepackage{booktabs}       
\usepackage{amsfonts}       
\usepackage{nicefrac}       
\usepackage{microtype}      
\usepackage{xcolor}         

\usepackage{graphicx}
\usepackage{wrapfig}
\usepackage{multirow}

\title{Understanding Failure Patterns in LLM-Based Mathematical Optimization Modeling}

\author{%
  Ryoichi Takase$^{1}$ \quad Xian Yeow Lee$^{1}$ \quad Hikaru Tomonari$^{2}$ \quad Masaya Tsunokake$^{2}$ \\
  $^{1}$Hitachi America, Ltd., California, USA\\
  $^{2}$Hitachi, Ltd., Tokyo, Japan\\
  \texttt{\{ryoichi.takase, xian.lee\}@hal.hitachi.com} \\
  \texttt{\{hikaru.tomonari.oj, masaya.tsunokake.qu\}@hitachi.com} \\
}

\begin{document}

\maketitle

\begin{abstract}
Industrial optimization involves complex constraints, making optimization modeling difficult to automate. Large language models (LLMs) have been explored for generating optimization models from natural language, but their reliability in end-to-end settings remains limited, often producing non-executable, infeasible, or feasible but incorrect solutions. In this work, we study failure modes in LLM-based optimization modeling by decomposing the process into three components: mathematical reasoning, solver usage, and domain knowledge, and constructing datasets that target each aspect. Using these datasets, we analyze how different components relate to errors and find that they are associated with different types of failure, although these relationships are not always strong or consistent. Furthermore, while combining these components improves execution reliability, feasible but incorrect solutions remain the dominant failure mode, suggesting a remaining gap between executability and correctness. These findings highlight limitations of current dataset designs and suggest the importance of data that better capture constraint completeness and formulation accuracy to enhance reliability.
\end{abstract}

\section{Introduction}
Optimization is central to industrial operations, where planning and scheduling tasks must satisfy complex constraints arising from safety requirements and regulatory compliance. Optimization frameworks such as linear programming (LP) and mixed‑integer linear programming (MILP) remain fundamental for achieving efficient and reliable solutions in industrial applications~\cite{wikarek2025proactive,kantor2020mixed}. Meanwhile, constructing such models typically requires domain expertise and manual effort, limiting scalability and adaptability to new operational scenarios.

Recent progress in large language models (LLMs) has motivated efforts to automatically generate optimization models from natural-language descriptions~\cite{ramamonjison2022nl4opt,xiao2024chain,zhuohan2025optibench}. While these approaches demonstrate promising results, achieving high reliability in industrial-scale problems remains challenging. In practice, LLM-based optimization systems often fail in multiple ways, including generating non-executable solver code, producing infeasible formulations, or yielding feasible but incorrect solutions. Existing work has primarily focused on improving modeling accuracy, with limited attention to analyzing failure modes in LLM-generated outputs.

In this work, we analyze failure modes in LLM-based optimization modeling. We decompose the modeling process into three components---mathematical reasoning, solver usage, and domain knowledge---and construct datasets that explicitly capture each component. Using these datasets, we examine how different data elements relate to errors in end-to-end generation. Our analysis shows that different components are associated with distinct failure types, such as non-executable code and infeasible formulations. While integrating these components improves execution reliability, it does not substantially reduce \textit{feasible but incorrect} solutions, which remain the dominant failure mode across benchmarks, suggesting a gap between executability and correctness.

\section{Method}
\subsection{End-to-End Optimization Modeling}
We define an end-to-end pipeline that maps natural-language descriptions to optimization solutions. The LLM generates solver-compatible code, which is then executed by a solver to obtain a solution. The LLM is responsible for (i) translating problem descriptions into mathematical formulations, (ii) implementing these formulations as executable solver code, and (iii) incorporating domain-specific constraints required for realistic settings. Errors may arise from multiple sources, including formulation and solver implementation, and are often entangled in the generated code, making it difficult to produce correct solutions.

\subsection{Decomposing Optimization Modeling into Core Components}
Building on the role of LLM in end-to-end optimization modeling, we decompose the process into three key elements that define distinct aspects of the dataset:
\begin{itemize}
\item \textbf{Mathematical reasoning:} Translating real-world requirements into optimization models involves a structured derivation from informal specifications to decision variables, objective functions, and constraints. Data that captures such reasoning processes enables models to learn how to construct well-formed mathematical representations.

\item \textbf{Solver usage:} Constructing executable optimization models requires knowledge of solver-specific APIs and configuration details (e.g., variable types, linearity constraints, and syntax) as in PySCIPOpt~\cite{maher2016pyscipopt}. Data reflecting solver interaction—including both correct usage and common implementation errors—supports the generation of executable and debuggable code.

\item \textbf{Domain knowledge:} Realistic optimization problems depend on domain-specific constraints and operational rules. Data incorporating such knowledge defines feasible regions and dependencies, helping prevent invalid modeling assumptions and improving solution validity.
\end{itemize}

These elements are tightly coupled, and their interactions may give rise to distinct failure modes. This design allows us to examine how different aspects of the dataset relate to execution reliability and solution correctness. We hypothesize that each component is associated with specific failure modes and that its coverage may influence overall model reliability.

\subsection{Constructing Datasets via LLM and Solver Feedback}
\begin{figure*}[t]
\centering
\includegraphics[width=\linewidth]{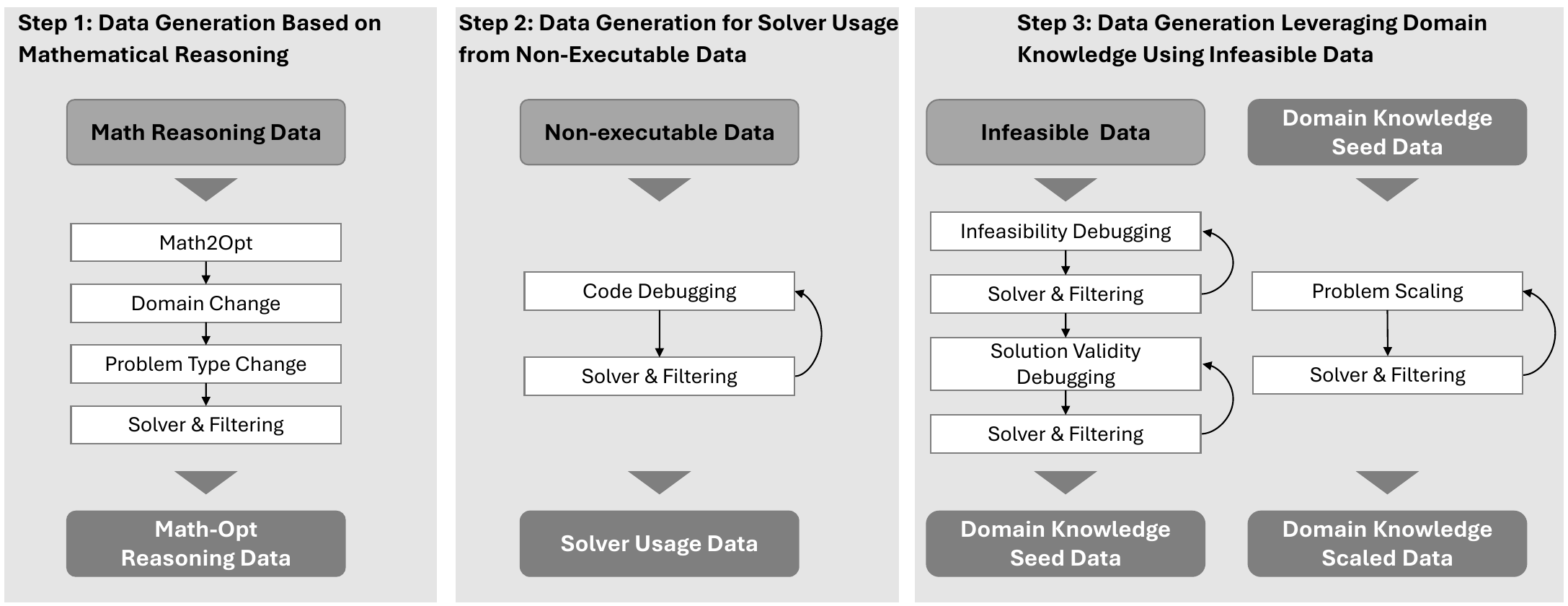}
\caption{Overview of the three-step data construction pipeline.}
\label{fig:data_generation_overview}
\end{figure*}

Figure~\ref{fig:data_generation_overview} provides an overview of the data generation pipeline. Datasets are constructed using an LLM-driven framework that generates optimization modeling tasks through prompting and solver-based feedback. \textbf{Math-Opt Reasoning} data are generated by prompting the LLM to produce optimization problems in the form of natural-language descriptions, mathematical formulations, and solver code, followed by solver-based filtering to retain only solvable instances. \textbf{Solver Usage} data are constructed by applying iterative debugging to non-executable outputs using error messages and code correction prompts, capturing common issues in implementation and solver execution. \textbf{Domain Knowledge} data are obtained by refining infeasible problems with domain-specific constraints and expanding them through iterative scaling, resulting in more realistic and complex optimization tasks.

In this study, we use gpt‑oss‑120b~\cite{agarwal2025gpt} as the data-generating LLM and the s1‑1.1 dataset~\cite{muennighoff2025s1} as the starting point for data construction. This process yields three datasets---Math-Opt Reasoning (32K samples), Solver Usage (4K samples), and Domain Knowledge (8K samples). For each dataset, 5\%--8\% of samples are reserved for testing and the remainder for training and validation.

\section{Results}
\subsection{Dataset Characteristics and Failure Analysis}
\begin{figure*}[!!b]
  \begin{minipage}[b]{0.9\linewidth}
    \centering
    \includegraphics[keepaspectratio, scale=0.47]{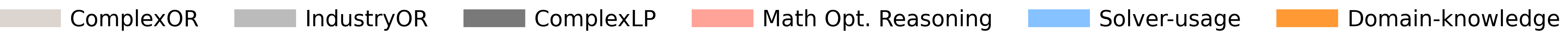}
  \end{minipage}
  \begin{minipage}[b]{0.24\linewidth}
    \centering
    \includegraphics[keepaspectratio, scale=0.3]{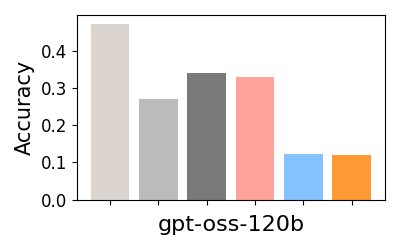}
  \end{minipage}
  \begin{minipage}[b]{0.24\linewidth}
    \centering
    \includegraphics[keepaspectratio, scale=0.3]{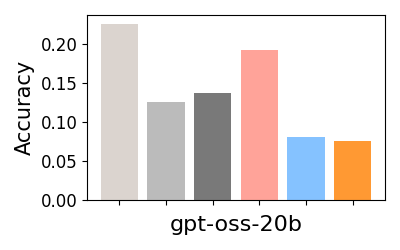}
  \end{minipage}
  \begin{minipage}[b]{0.24\linewidth}
    \centering
    \includegraphics[keepaspectratio, scale=0.3]{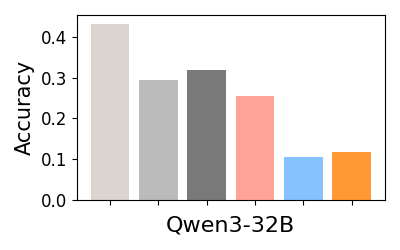}
  \end{minipage}
  \begin{minipage}[b]{0.24\linewidth}
    \centering
    \includegraphics[keepaspectratio, scale=0.3]{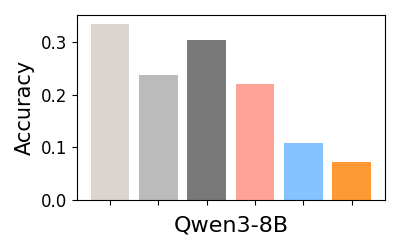}
  \end{minipage}\\

  \begin{minipage}[b]{0.24\linewidth}
    \centering
    \includegraphics[keepaspectratio, scale=0.3]{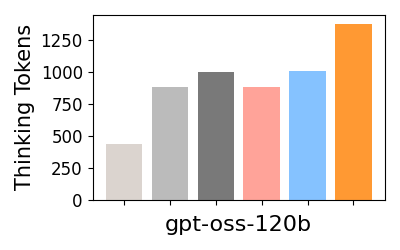}
  \end{minipage}
  \begin{minipage}[b]{0.24\linewidth}
    \centering
    \includegraphics[keepaspectratio, scale=0.3]{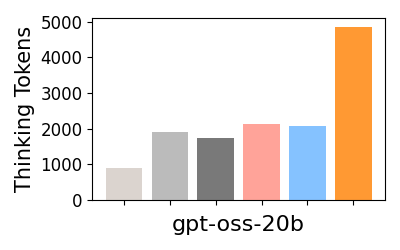}
  \end{minipage}
  \begin{minipage}[b]{0.24\linewidth}
    \centering
    \includegraphics[keepaspectratio, scale=0.3]{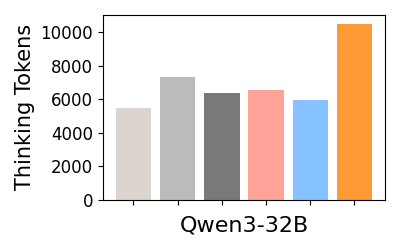}
  \end{minipage}
  \begin{minipage}[b]{0.24\linewidth}
    \centering
    \includegraphics[keepaspectratio, scale=0.3]{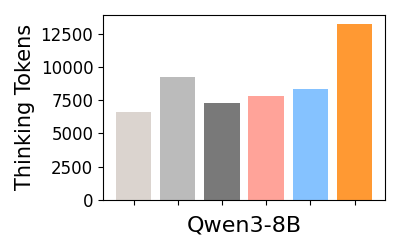}
  \end{minipage}

  \caption{Modeling accuracy and reasoning-token usage. The Math-Opt Reasoning dataset requires reasoning tokens comparable to existing benchmarks, while Solver Usage achieves lower accuracy with similar token lengths. Domain Knowledge requires more tokens due to domain-specific formulation complexity.}
  \label{fig:accuracy_thinking_token}
\end{figure*}

\begin{figure*}[!!t]
  \begin{minipage}[b]{0.9\linewidth}
    \centering
    \includegraphics[keepaspectratio, scale=0.47]{figs/data_analysis/data_analysis_label.png}
  \end{minipage}
  \begin{minipage}[b]{0.24\linewidth}
    \centering
    \includegraphics[keepaspectratio, scale=0.3]{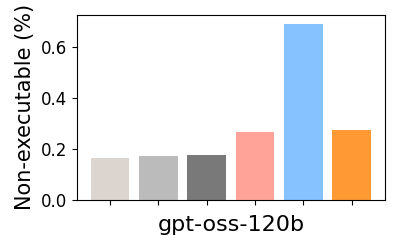}
  \end{minipage}
  \begin{minipage}[b]{0.24\linewidth}
    \centering
    \includegraphics[keepaspectratio, scale=0.3]{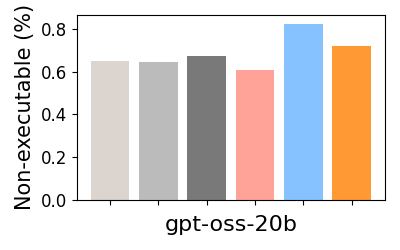}
  \end{minipage}
  \begin{minipage}[b]{0.24\linewidth}
    \centering
    \includegraphics[keepaspectratio, scale=0.3]{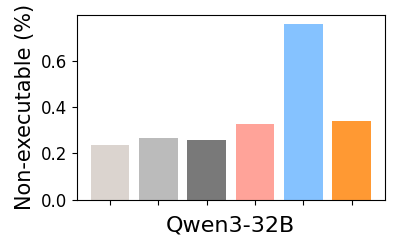}
  \end{minipage}
  \begin{minipage}[b]{0.24\linewidth}
    \centering
    \includegraphics[keepaspectratio, scale=0.3]{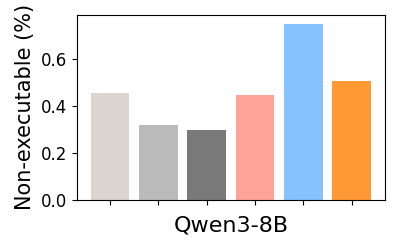}
  \end{minipage}\\

  \begin{minipage}[b]{0.24\linewidth}
    \centering
    \includegraphics[keepaspectratio, scale=0.3]{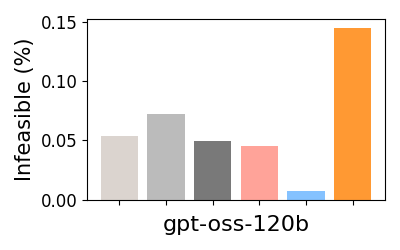}
  \end{minipage}
  \begin{minipage}[b]{0.24\linewidth}
    \centering
    \includegraphics[keepaspectratio, scale=0.3]{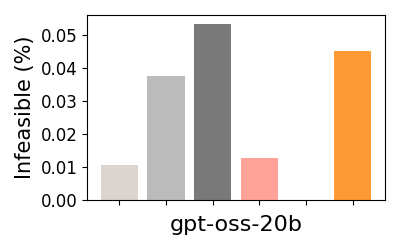}
  \end{minipage}
  \begin{minipage}[b]{0.24\linewidth}
    \centering
    \includegraphics[keepaspectratio, scale=0.3]{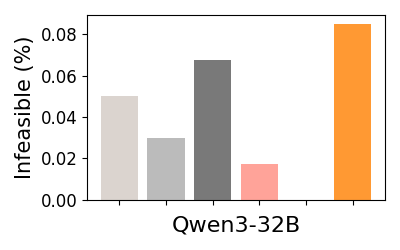}
  \end{minipage}
  \begin{minipage}[b]{0.24\linewidth}
    \centering
    \includegraphics[keepaspectratio, scale=0.3]{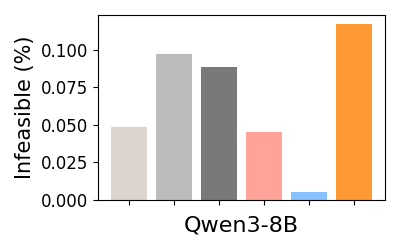}
  \end{minipage}

  \caption{Non-executable and infeasible rates. Math-Opt Reasoning shows error rates comparable to existing benchmarks, while Solver Usage has higher non-executable rates and Domain Knowledge higher infeasible rates, highlighting distinct challenges.}
  \label{fig:non_executable_infeasibility}
\end{figure*}

To enable comparison of how different datasets capture distinct aspects of optimization modeling, we examine the model responses using off‑the‑shelf LLMs (gpt‑oss 120B/20B and Qwen3 32B/8B). These models are used only as diagnostic tools to reveal dataset properties, and the reported metrics are not intended as model performance benchmarks. In Figures~\ref{fig:accuracy_thinking_token} and~\ref{fig:non_executable_infeasibility}, the Math‑Opt Reasoning, Solver Usage, and Domain Knowledge datasets are shown in red, blue, and orange, respectively, while existing benchmarks such as ComplexOR~\cite{xiao2024chain}, IndustryOR~\cite{huang2024orlm}, and ComplexLP~\cite{xuhan2024mamo} are shown in grayscale.

As shown in Figure~\ref{fig:accuracy_thinking_token}, the Math‑Opt Reasoning dataset has reasoning lengths comparable to existing benchmarks. Solver Usage shows similar lengths but lower accuracy, indicating challenges in generating valid solver code. Domain Knowledge requires more reasoning tokens yet achieves lower accuracy, reflecting the complexity of incorporating domain-specific constraints. These results suggest that dataset difficulty depends not only on reasoning length but also on solver requirements and constraint complexity.

Figure~\ref{fig:non_executable_infeasibility} compares non-executable and infeasible rates, revealing distinct failure modes. Solver Usage has the highest non-executable rate, while Domain Knowledge produces more infeasible solutions due to complex constraints. In contrast, Math‑Opt Reasoning shows error rates comparable to existing benchmarks. Overall, non-executable errors and infeasibility arise from different sources, reflecting distinct challenges in optimization modeling.

\subsection{Effect and Limitations of Data Components}
To analyze the contribution of each data component, we fine-tune LLMs on the constructed datasets and perform ablation studies by removing reasoning, solver usage, or domain knowledge. We evaluate models on standard benchmarks (ComplexOR, IndustryOR, ComplexLP, NLP4LP, NL4Opt), categorizing outputs into five types: non-executable, timeout, infeasible, feasible but incorrect, and correct. To account for stochasticity, we sample multiple outputs per instance (32 for ComplexOR and 4 for others) and report averaged results.

Experiments on Qwen3 models (1.7B and 4B)~\cite{yang2025qwen3}, using teacher data generated from a Qwen3-32B model via solver-based rejection sampling, show that the full dataset consistently yields the best performance (Tables \ref{tab:ablation_study_1B} and \ref{tab:ablation_study_4B}), suggesting that all components are beneficial. Removing solver usage data increases non-executable outputs, although the causal relationship remains unclear.

\begin{figure*}[!!t]
  \begin{minipage}[b]{0.3\linewidth}
    \centering
    \includegraphics[keepaspectratio, scale=0.3]{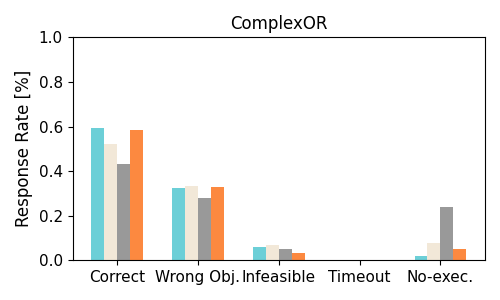}
  \end{minipage}
  \begin{minipage}[b]{0.3\linewidth}
    \centering
    \includegraphics[keepaspectratio, scale=0.3]{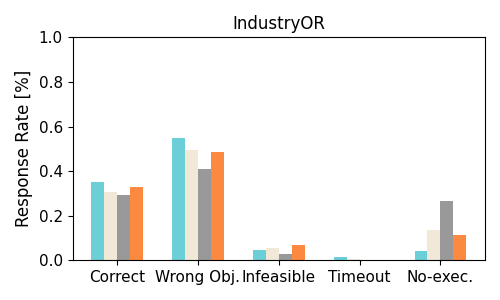}
  \end{minipage}
  \begin{minipage}[b]{0.3\linewidth}
    \centering
    \includegraphics[keepaspectratio, scale=0.3]{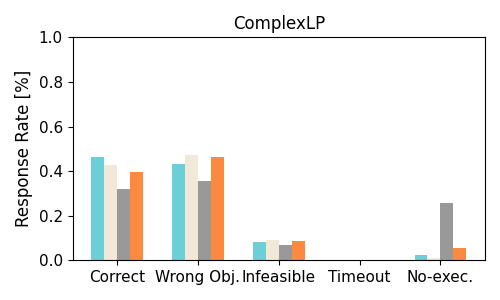}
  \end{minipage}\\
  \begin{minipage}[b]{0.3\linewidth}
    \centering
    \includegraphics[keepaspectratio, scale=0.3]{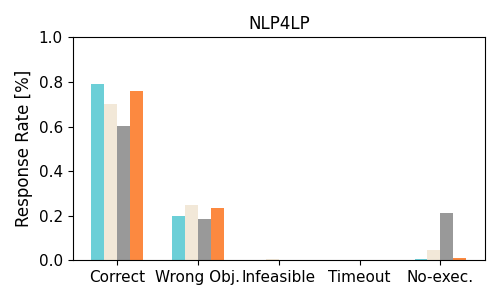}
  \end{minipage}
  \begin{minipage}[b]{0.3\linewidth}
    \centering
    \includegraphics[keepaspectratio, scale=0.3]{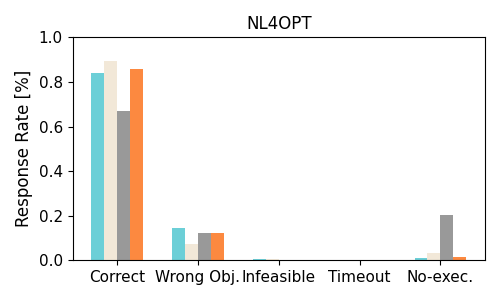}
  \end{minipage}
  \begin{minipage}[b]{0.3\linewidth}
    \centering
    \includegraphics[keepaspectratio, scale=0.5]{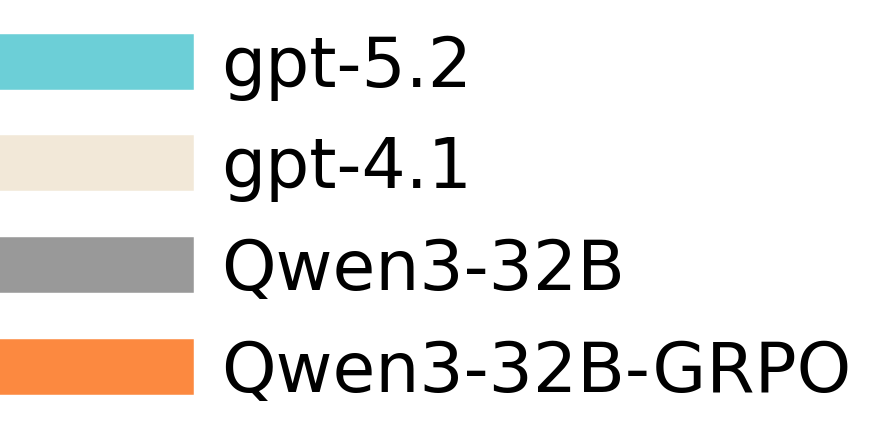}
  \end{minipage}

  \caption{Failure patterns across benchmarks: although the non-executable rate decreases with fine-tuning, LLMs still produce wrong objectives (i.e., feasible but incorrect solutions), particularly in more complex tasks.}
  \label{fig:response_category}
\end{figure*}

Nevertheless, across all settings, including larger 32B models fine-tuned with GRPO~\cite{shao2024deepseekmath}, feasible but incorrect outputs (denoted as Wrong Obj.) remain dominant (Figure~\ref{fig:response_category}). Although fine-tuning reduces non-executable errors, this failure mode persists and becomes more pronounced as task complexity increases, indicating that LLMs often generate plausible yet invalid optimization formulations.

Our results reveal a key limitation of current LLM-based optimization systems: while data-centric design reduces explicit failures such as non-executable outputs, it does not adequately address formulation errors. In particular, feasible but incorrect solutions remain dominant, often caused by subtle issues such as missing constraints or incorrect objectives that are difficult to detect through standard supervision or execution-based validation. This suggests that improving reliability requires datasets that explicitly capture formulation structure, constraint completeness, and boundary conditions, enabling models to distinguish correct formulations from near-correct ones.

\section{Conclusion}
This work presents an analysis of failure modes in LLM-based optimization modeling. We show that different data components correspond to distinct failure modes and that integrating them improves execution reliability. However, feasible but incorrect solutions remain the dominant failure mode, highlighting a gap between executability and correctness. Addressing this limitation requires new dataset designs that better capture the structure of valid optimization formulations and reduce such errors in end-to-end pipelines.

{
\small
\bibliographystyle{plain}
\bibliography{ref}
}

\appendix

\section{Method Details}
\subsection{Three Key Elements in Mathematical Optimization}

Formulating real-world problems as solvable optimization models requires integrating three key elements to ensure correctness and practical feasibility.
\begin{itemize}
\item \textbf{Mathematical reasoning:} Transforming real-world requirements into optimization models requires a logical, step-by-step derivation from informal specifications to decision variables, objectives, and constraints. This reasoning ensures that optimization models capture the essential structure of the problem and yield well-defined mathematical formulations.

\item \textbf{Solver usage:} Implementing optimization models correctly requires familiarity with solver-specific APIs and parameters (e.g., variable types and linearity/convexity requirements) as in Gurobi~\cite{gurobi} and PySCIPOpt~\cite{maher2016pyscipopt}. Understanding syntax, configuration, and debugging is essential for producing executable, solver-ready code.

\item \textbf{Domain knowledge:} Background knowledge of operational rules and industry constraints in OT domains helps define the feasible decision space and its dependencies, thereby preventing invalid modeling assumptions and ensuring that optimization formulations maintain correctness and practical feasibility.
\end{itemize}

Motivated by the importance of combining mathematical reasoning, appropriate solver usage, and industry-specific domain knowledge, we propose a data‑generation method that integrates these three elements to train LLMs to produce solvable optimization models. As shown in Figure~\ref{fig:data_generation_overview}, our pipeline unifies these key components, which are described in detail below.

\subsection{Step 1: Data Generation Based on Mathematical Reasoning}
This data-generation step extracts multi-step reasoning traces from mathematical reasoning datasets and converts them into optimization modeling tasks covering a wide range of domains and problem types. Solver-based filtering retains only solvable problems that require both reasoning and modeling skills.

\textbf{\textit{Math2Opt}} first prompts an LLM to generate a multi-step reasoning trace from a math problem and its solution, and then converts this trace into a new mathematical optimization problem. The prompt instructs the LLM to construct an optimization model whose variables, constraints, and objective are derived from the reasoning steps. The resulting data consists of three components: (i) a natural‑language problem description, (ii) a mathematical formulation, and (iii) PySCIPOpt code. This process ensures that the generated data imposes multi-step reasoning on LLMs rather than simple text‑to‑equation conversion.

\textbf{\textit{Domain Change}} increases dataset diversity by mapping each seed optimization problem to one of 18 application domains (e.g., manufacturing, supply chain, energy), following~\cite{ahmaditeshnizi2023optimus,lima2025toward}. For each seed problem, an LLM generates a new version in a specified target domain while preserving the structural pattern of the original formulation. By systematically creating domain‑specific variations, this process enhances cross‑domain reasoning in LLMs.

\textbf{\textit{Problem Type Change}} further increases dataset diversity by introducing multiple classes of optimization problems, including Linear Programming (LP), Nonlinear Programming (NLP), Quadratic Programming (QP), and Mixed-Integer Linear Programming (MILP). Similar to the previous step, the LLM reformulates each seed problem into a new problem belonging to one of these classes, enabling LLMs to deal with a wide range of objectives and constraints.

\textbf{\textit{Solver \& Filtering}} validates the generated problems by executing solvers and retaining only those that yield optimal solutions. These filtered samples constitute the \textit{Math‑Opt Reasoning Data}. Samples that fail to solve are used for subsequent steps.

\subsection{Step 2: Data Generation for Solver Usage from Non-Executable Data}
Among the failed samples from Step~1, we categorize the errors as follows:
\begin{itemize}
\item \textbf{Infeasible}: No solution exists under the given objectives and constraints.
\item \textbf{Timeout}: Computation could not be completed within the specified time limit.
\item \textbf{Non-executable}: Solver execution failed due to errors.
\end{itemize}

Infeasible samples typically require substantial reformulation of the objectives or constraints and are therefore used in Step 3, which leverages domain knowledge. Timeout samples constitute only a small fraction of failures and are excluded from data generation.

In Step~2, we focus on non-executable samples and debug them to extract the cases from which the correct solver code can be successfully recovered. These samples are typically resolved by fixing code or configuration errors, and the resulting corrections provide valuable information on the appropriate problem formulations.

\textbf{\textit{Code Debugging}} is applied to non-executable code in representative OT domains, such as manufacturing, supply chain, and energy, to reduce data construction time. These domains provide diverse formulations and solver usage, ensuring sufficient dataset coverage. A debugging prompt is constructed by combining the problem description, source code, and associated error messages, and is then provided to the LLM. By resolving common errors that prevent successful execution, this process yields datasets that help LLMs learn the correct optimization formulation and solver configuration. Only code that becomes executable after several debugging iterations is reserved as \textit{Solver Usage Data}. Samples that remain non-executable or result in timeouts after the debugging iterations are discarded, whereas infeasible samples are used in Step~3.

\subsection{Step 3: Data Generation Leveraging Domain Knowledge Using Infeasible Data}

In real-world optimization problems, operational rules and requirements often introduce complex constraints that can produce infeasible solutions. In Step~3, the infeasible samples obtained from the previous steps are iteratively analyzed and corrected, thus ensuring the validity of the resulting dataset. Furthermore, by scaling the problem size, we construct more domain-aware optimization problems that integrate relevant expert knowledge.

\textbf{\textit{Infeasibility Debugging}} refines infeasible samples by prompting the LLM to determine whether the infeasibility arises from a coding error or a modeling issue, then minimally reformulates the problem and updates the code while preserving intent. By making only minimal changes to restore feasibility, the method produces samples near the feasible--infeasible boundary.

\textbf{\textit{Solution Validity Debugging}} constructs a prompt from the problem description, code, and solver output to obtain relevant domain knowledge from the LLM\footnote{In this study, domain knowledge refers to short, listed statements describing physical or operational rules that are generated by the LLM. For example, in the energy domain, the generated list includes statements on charger efficiency, grid peak‑demand limits, and tariff constraints.}. This domain knowledge is then used to revise the objective and constraints, so the optimization model yields realistic solutions. During iteration, samples that time out or become non-executable are processed again to obtain feasible solutions, while samples that become infeasible are handled with infeasibility debugging. In the end, this process produces the \textit{Domain Knowledge Seed Data}, which serve as starting points for scaling the optimization problem.

\textbf{\textit{Problem Scaling}} expands the Domain Knowledge Seed Data by applying one of the three types of domain knowledge, such as \textit{insights}, \textit{operational rules}, or \textit{tacit knowledge}. Each type is used to introduce additional decision variables or constraints. For example, when leveraging operational rules, we scale the problem by adding new decision variables in the target OT domain using the following prompt: \textit{``In particular, incorporate key operational rules from the energy domain, increase the number of decision variables, and significantly expand the problem scale''}\footnote{Examples of added operational rules include required shut‑down periods for thermal protection and associated start‑up cost effects.}. 

After an optimal solution is found under the initial scaling, additional scaling is applied to further expand the problem. If this scaling results in timeouts or non-executable samples, the previous scaling level is retried to obtain an optimal solution. When infeasibility arises, infeasibility debugging with domain knowledge is applied. Samples that yield optimal solutions after this process are reserved as \textit{Domain Knowledge Scaled Data}.

In the proposed method, we used the gpt‑oss‑120b~\cite{agarwal2025gpt} as the data‑generating LLM and the s1‑1.1 dataset~\cite{muennighoff2025s1} as the starting point for mathematical reasoning data, from which we produced four datasets: Math‑Opt Reasoning Data (31,773 samples), Solver Usage Data (4,264 samples), Domain Knowledge Seed Data (735 samples), and Domain Knowledge Scaled Data (7,009 samples). For each dataset, approximately 5\% to 8\% of the samples were reserved as test data, while the remaining samples were used for training and validation purposes.

\subsection{Training Reasoning LLMs}
\begin{figure*}[t]
\centering
\includegraphics[width=\linewidth]{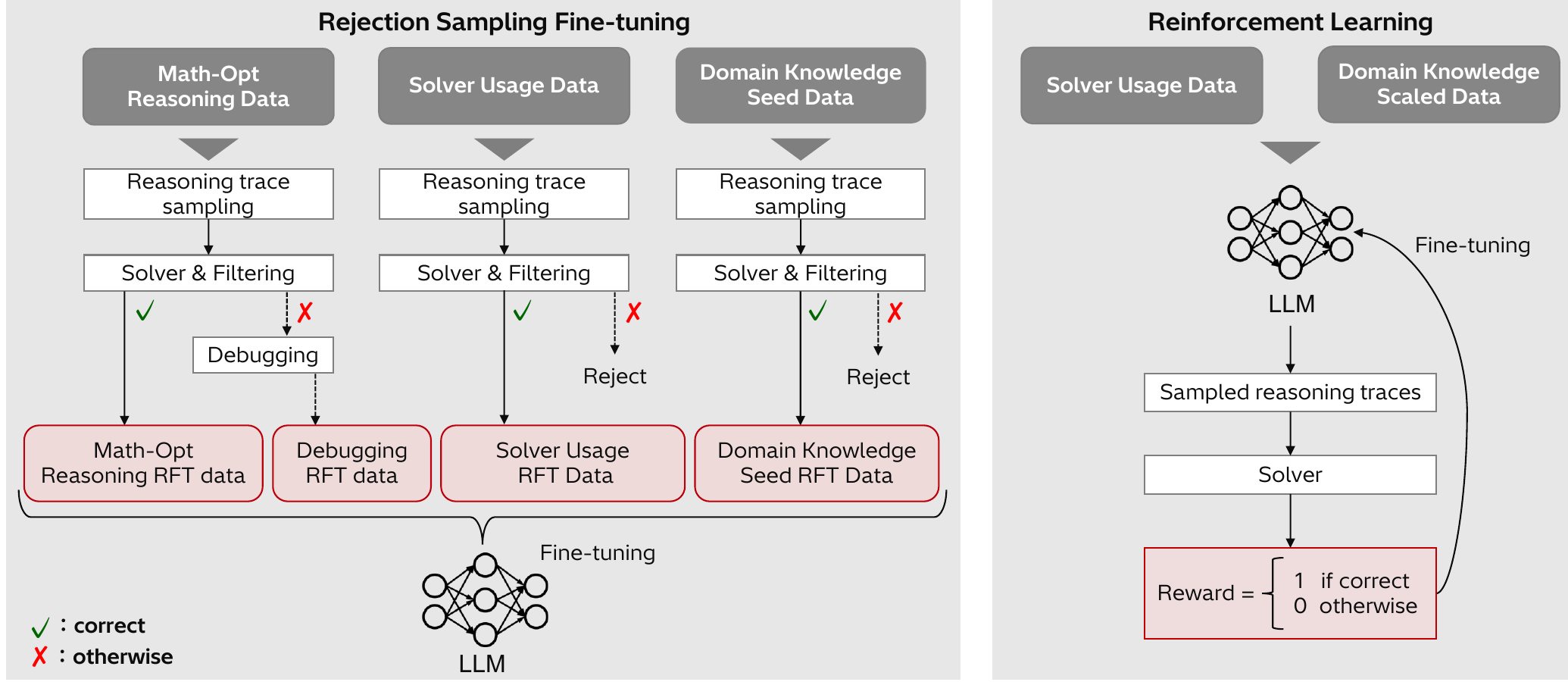}
\caption{Overview of training reasoning LLMs. Left: Rejection Sampling Fine‑Tuning (RFT), which generates multiple reasoning traces using Math‑Opt Reasoning Data, Solver Usage Data, and Domain Knowledge Seed Data, using only solver‑verified correct traces for training. Right: Reinforcement Learning (RL), which trains the model on Solver Usage Data and Domain Knowledge Scaled Data, with solver‑executed reasoning traces providing binary rewards based on objective value correctness.}
\label{fig:training_overview}
\vspace{-3mm}
\end{figure*}

In this study, we adopt Qwen3-32B~\cite{yang2025qwen3} as the base reasoning model. 
Our dataset contains the problem description, code, and ground‑truth objective value, but does not include the reasoning traces required for training. To obtain these traces, we apply Qwen3‑32B to our dataset and generate the corresponding reasoning traces. As shown in Figure~\ref{fig:training_overview}, we first apply Rejection Sampling Fine-Tuning (RFT) to improve the accuracy of mathematical optimization modeling, followed by reinforcement learning (RL) to further enhance overall performance.

\subsubsection*{Rejection Sampling Fine-Tuning (RFT)}
We construct training data using Math-Opt Reasoning Data, Solver Usage Data, and Domain Knowledge Seed Data. For each problem, we sample 8 reasoning traces from the Math-Opt Reasoning Data and 16 reasoning paths from the Solver Usage Data and Domain Knowledge Seed Data. We then execute a solver on the sampled reasoning paths and extract only those whose objective values match the ground‑truth results, thereby producing three types of RFT data: Math-Opt Reasoning RFT Data (57,025 samples), Solver Usage RFT Data (1,477 samples), and Domain Knowledge Seed RFT Data (946 samples)\footnote{Solver Usage Data and Domain Knowledge Seed Data are more difficult than Math-Opt Reasoning Data. The pass rates for Math-Opt Reasoning Data, Solver Usage Data, and Domain Knowledge Seed Data are 0.27, 0.051, and 0.17, respectively. In addition, RFT for Solver Usage Data and Domain Knowledge Seed Data use only subsets that are not included in RL training, resulting in fewer correct samples compared to Math-Opt Reasoning Data.
}.

For the Math-Opt Reasoning Data, we additionally perform debugging, and any correctly solved samples are incorporated into the Debugging RFT Data (24,253 samples). In contrast, debugging is not performed for the Solver Usage Data or the Domain Knowledge Seed Data, as these will be repeatedly evaluated later through RL.

\subsubsection*{Reinforcement Learning (RL)}
We perform Group Relative Policy Optimization (GRPO)~\cite{shao2024deepseekmath} using the Solver Usage Data and the Domain Knowledge Scaled Data. To compute the reward, we first extract the code from the reasoning traces generated by the LLM, and then execute a solver on the extracted code to obtain its objective value. We assign a reward of 1 if the resulting objective value matches the ground‑truth value, and 0 otherwise. This reward function is designed to directly reflect optimization‑modeling quality by evaluating the accuracy of the objective values. The LLM parameters are then updated to maximize this reward, thereby improving the LLM’s mathematical optimization modeling capability. In this study, we set the sampling parameters to temperature = 0.6, TopK = 20, and TopP = 0.95. For each problem, four reasoning trajectories were generated. The RL training is performed using verl \cite{sheng2024hybridflow}.

\section{Additional Results}
\subsection{Ablation Study on Small Models}
To verify that integrating mathematical reasoning, solver usage, and domain knowledge leads to improved performance, we conduct an ablation study with four settings: using all three components, removing Math-Opt Reasoning data, removing Solver Usage data, and removing Domain Knowledge data. Each dataset remains identical across configurations, and the total number of training steps and the overall amount of data used to update the LLM are kept constant across all four cases. We conducted the ablations on the 1.7B and 4B models due to the computational cost of 32B‑scale experiments, while maintaining sensitivity to performance differences. The results are shown in Tables~\ref{tab:ablation_study_1B} and \ref{tab:ablation_study_4B}. The Full setting achieves the best overall performance, indicating that combining all three forms of supervision is crucial. Although further analysis is needed to quantify each component's contribution, the consistent trend across both the 1.7B and 4B models supports the benefit of integrating mathematical reasoning, solver usage, and domain knowledge.

{
\tabcolsep = 8pt

\begin{table*}[t]
\centering
\caption{Ablation study results for Qwen3-1.7B on the full dataset and variants without mathematical reasoning, solver usage, or domain knowledge.}

\resizebox{\textwidth}{!}{%
\begin{tabular}{l|ccccccccccccccccc}\toprule
Dataset Variant (Qwen3-1.7B)	&	Correct	&	Wrong Obj.	&	Infeasible	&	Timeout	&	Non-exec.	\\\midrule
Full	&	0.394	&	0.268	&	0.033	&	0.000	&	0.305	\\
W/O Math Opt. Reasoning	&	0.362	&	0.310	&	0.057	&	0.001	&	0.271	\\
W/O Solver Usage	&	0.376	&	0.249	&	0.032	&	0.000	&	0.343	\\
W/O Domain Knowledge	&	0.357	&	0.282	&	0.037	&	0.000	&	0.325	\\\midrule
\end{tabular}
}

\label{tab:ablation_study_1B}
\end{table*}
}

{
\tabcolsep = 8pt

\begin{table*}[t]
\centering
\caption{Ablation study results for Qwen3-4B on the full dataset and variants without mathematical reasoning, solver usage, or domain knowledge.}

\resizebox{\textwidth}{!}{%
\begin{tabular}{l|ccccccccccccccccc}\toprule
Dataset Variant (Qwen3-4B)	&	Correct	&	Wrong Obj.	&	Infeasible	&	Timeout	&	Non-exec.	\\\midrule
Full	&	0.503	&	0.280	&	0.044	&	0.001	&	0.173	\\
W/O Math Opt. Reasoning	&	0.500	&	0.278	&	0.051	&	0.001	&	0.169	\\
W/O Solver Usage	&	0.459	&	0.294	&	0.055	&	0.000	&	0.192	\\
W/O Domain Knowledge	&	0.497	&	0.304	&	0.042	&	0.001	&	0.157	\\\midrule
\end{tabular}
}

\label{tab:ablation_study_4B}
\end{table*}
}

\subsection{Reasoning LLM Training Results}
For performance evaluation, we adopted five benchmark datasets: ComplexOR~\cite{xiao2024chain}, IndustryOR~\cite{huang2024orlm}, NL4Opt~\cite{ramamonjison2022nl4opt}, NLP4LP~\cite{ahmaditeshnizi2023optimus}, and ComplexLP~\cite{xuhan2024mamo}, which are widely used in optimization modeling research \cite{zhuohan2025optibench}. 
To ensure the robustness of the results, we employed a multi-sampling strategy. Specifically, for the ComplexOR dataset, 32 candidate solutions were sampled for each problem due to the small amout of data size, while for the remaining datasets, four candidate solutions per problem were generated. The final accuracy was computed as the average across all sampled solutions. This approach mitigates variance caused by stochastic generation and provides a more reliable estimate of model performance.

Table~\ref{tab:performance-comparison} shows the training results. Supervised fine-tuning and reinforcement learning consistently improve optimization-modeling performance over the base model, and the 32B model reaches performance comparable to or exceeding o3 on several benchmarks. However, the key observation is that improved benchmark performance does not fully resolve the reliability challenges of optimization modeling. The dominant remaining failure mode is \textit{feasible but incorrect} solutions, indicating that many generated formulations can be executed successfully while still encoding incorrect objectives or constraints. This finding highlights the importance of datasets and training signals that emphasize formulation correctness in addition to executability.

{
\tabcolsep = 8pt

\begin{table*}[h]
\centering
\caption{Dataset‑level accuracy comparison across all models on the benchmarks.}

\resizebox{\textwidth}{!}{%
\begin{tabular}{l|ccccc|cccccccccccc}\toprule
Model	&	ComplexOR	&	IndustryOR	&	ComplexLP	&	NLP4LP	&	NL4OPT	&	Average	\\\midrule
GPT-5.2	&	0.595$\pm$0.046	&	0.353$\pm$0.037	&	0.463$\pm$0.080	&	0.791$\pm$0.053	&	0.840$\pm$0.037	&	0.609$\pm$0.050	\\
o3	&	0.552$\pm$0.159	&	0.330$\pm$0.119	&	0.392$\pm$0.181	&	0.748$\pm$0.070	&	0.886$\pm$0.074	&	0.582$\pm$0.121	\\
GPT-4.1	&	0.521$\pm$0.100	&	0.308$\pm$0.082	&	0.427$\pm$0.117	&	0.700$\pm$0.068	&	0.892$\pm$0.063	&	0.570$\pm$0.086	\\\midrule
gpt-oss-120b	&	0.472$\pm$0.227	&	0.270$\pm$0.113	&	0.341$\pm$0.207	&	0.590$\pm$0.190	&	0.692$\pm$0.253	&	0.473$\pm$0.198	\\
gpt-oss-20b	&	0.226$\pm$0.247	&	0.125$\pm$0.142	&	0.137$\pm$0.185	&	0.374$\pm$0.313	&	0.460$\pm$0.400	&	0.264$\pm$0.257	\\
Qwen3-32B	&	0.432$\pm$0.224	&	0.295$\pm$0.120	&	0.318$\pm$0.198	&	0.602$\pm$0.262	&	0.672$\pm$0.292	&	0.464$\pm$0.219	\\
Qwen3-8B	&	0.335$\pm$0.131	&	0.238$\pm$0.062	&	0.303$\pm$0.181	&	0.536$\pm$0.250	&	0.611$\pm$0.258	&	0.405$\pm$0.177	\\\midrule
Qwen3-32B-RFT	&	0.446$\pm$0.057	&	0.328$\pm$0.067	&	0.352$\pm$0.138	&	0.675$\pm$0.128	&	0.778$\pm$0.145	&	0.516$\pm$0.107	\\
Qwen3-32B-GRPO	&	0.583$\pm$0.038	&	0.330$\pm$0.026	&	0.395$\pm$0.026	&	0.757$\pm$0.032	&	0.860$\pm$0.032	&	0.585$\pm$0.031	\\\bottomrule

\end{tabular}
}

\label{tab:performance-comparison}
\end{table*}
}

\end{document}